\documentclass[11pt]{article}

\usepackage[preprint]{acl}
 \usepackage{booktabs} 
\usepackage{times}
\usepackage{latexsym}

\usepackage[T1]{fontenc}

\usepackage[utf8]{inputenc}

\usepackage{microtype}

\usepackage{inconsolata}

\usepackage{graphicx}

\title{Think$^{2}$: Grounded Metacognitive Reasoning in Large Language Models}

\author{
 \textbf{Abraham Paul Elenjical\textsuperscript{1}\thanks{These authors contributed equally.}}, 
 \textbf{Vivek Hruday Kavuri\textsuperscript{1}\footnotemark[1]}, 
 \textbf{Vasudeva Varma\textsuperscript{1}}
\\
 \textsuperscript{1}IIIT Hyderabad,
\\
 \small{
   \textbf{Correspondence:}
   \texttt{\{abraham.elenjical, kavuri.hruday\}@research.iiit.ac.in}
 }
}

\begin{document}
\maketitle

\begin{abstract}
Large Language Models (LLMs) demonstrate strong reasoning performance, yet their ability to reliably monitor, diagnose, and correct their own errors remains limited. We introduce a psychologically grounded metacognitive framework that operationalizes Ann Brown’s regulatory cycle (Planning, Monitoring, and Evaluation) as a structured prompting architecture, and study its integration within a lightweight dual-process MetaController for adaptive effort allocation. Across diverse reasoning and diagnostic benchmarks (GSM8K, CRUXEval, MBPP, AIME, CorrectBench, and TruthfulQA) using Llama-3 and Qwen-3 (8B), explicit regulatory structuring substantially improves error diagnosis and yields a threefold increase in successful self-correction. Blinded human evaluations over 580 query pairs show an 84\% aggregate preference for trustworthiness and metacognitive self-awareness over standard and Chain-of-Thought baselines. Grounding LLM reasoning in established cognitive theory offers a principled path toward more transparent and diagnostically robust AI systems.\footnote{Code will be released public upon acceptance.}
\end{abstract}

\section{Introduction}
\label{sec:introduction}

Large Language Models (LLMs) exhibit impressive reasoning abilities, yet their capacity for \emph{metacognitive regulation}, such as monitoring their reasoning, diagnosing errors, and inhibiting hallucinations, remains fragile. While Chain-of-Thought (CoT) prompting improves step-by-step reasoning \cite{wei2022chain}, it does not enforce structured self-regulation. Metacognitive Prompting (MP) \cite{wang-zhao-2024-metacognitive} encourages reflection, but largely treats metacognition as a heuristic generation strategy rather than a cognitively grounded process.

Educational psychology defines metacognition through three regulatory phases - \emph{Planning}, \emph{Monitoring}, and \emph{Evaluation} - that characterize expert problem solving \cite{Brown1987}. Yet, as shown by \citet{zhang-etal-2025-remembering}, many benchmarks used to evaluate metacognitive methods primarily assess lower-order skills such as \emph{Remembering} and \emph{Understanding}. Consequently, performance gains on standard NLU tasks may reflect verbosity rather than genuine regulatory competence.

We operationalize Ann Brown’s regulatory cycle \cite{Brown1987} as a structured prompting framework that explicitly separates Planning (strategy formation), Monitoring (controlled execution), and Evaluation (consistency checking). To measure regulatory behavior, we evaluate on tasks that inherently require higher-order control: GSM8K \cite{cobbe2021trainingverifierssolvemath}, CRUXEval \cite{gu2024cruxeval}, MBPP \cite{DBLP:journals/corr/abs-2108-07732}, AIME \cite{veeraboina_2024_aime}, CorrectBench \cite{talmor-etal-2019-commonsenseqa}, and TruthfulQA \cite{lin-etal-2022-truthfulqa}. These datasets stress sequential planning, execution monitoring, diagnostic correction, and deception resistance.

Our study addresses three research questions:

\begin{itemize}
    \item \textbf{RQ1 (Theoretical Efficacy):} Does grounding prompts in Ann Brown’s regulatory phases improve error detection and self-correction compared to heuristic metacognitive prompting?
        
    \item \textbf{RQ2 (Human Alignment):} Do psychologically grounded reasoning trajectories produce outputs perceived as more trustworthy and self-aware?

    \item \textbf{RQ3 (Cognitive Economy):} Can a lightweight dual-process \emph{MetaController}, inspired by adaptive effort allocation theories \cite{kahneman2011thinking}, route tasks between fast (System 1) and structured regulatory reasoning (System 2) without degrading performance?
\end{itemize}

Across Llama-3 and Qwen 8B models, we observe a key interaction between model capability and regulatory structure: uniformly enforced metacognitive prompting can impose overhead on instruction-tuned models but synergizes with native reasoning models. Blinded human evaluations over 580 query pairs further show consistent preference for psychologically grounded reasoning in trustworthiness and deployment suitability. Advancing LLM metacognition therefore requires structured regulatory control and evaluation protocols that genuinely measure it.

\section{Related Work}
\label{sec:related_work}

Our work connects metacognitive prompting, adaptive inference architectures, and cognitive benchmarking to argue for psychologically grounded and resource-aware reasoning.

\subsection{Metacognitive Prompting and Introspection}

Chain-of-Thought prompting \cite{wei2022chain} improved reasoning by eliciting intermediate steps but lacks explicit self-regulation. Metacognitive Prompting (MP) \cite{wang-zhao-2024-metacognitive} introduces a five-stage reflective process (comprehension, judgment, evaluation, decision, confidence), yet remains static and computationally uniform across tasks. Recent approaches address these limitations by either mapping to Schoenfeld's Episode Theory \cite{li-etal-2025-understanding}, or by enforcing explicit self-correction through programmatic verification \cite{song-etal-2025-progco}.

Related efforts include task decomposition \cite{wang2023planandsolvepromptingimprovingzeroshot}, iterative self-correction \cite{NEURIPS2023_91edff07}, and broader reasoning elicitation techniques \cite{huang2023reasoninglargelanguagemodels}. The Monitor-Generate-Verify framework \cite{oh-2025-before} operationalizes Flavell’s cognitive monitoring model, but emphasizes descriptive monitoring rather than prescriptive regulation. In contrast, our approach grounds regulation in Ann Brown’s functional metacognitive cycle \cite{Brown1987}, enabling explicit error diagnosis and remediation.

\subsection{Benchmarking Cognitive Depth}

Benchmark selection critically shapes conclusions about metacognition. \citet{zhang-etal-2025-remembering} show that widely used benchmarks such as MMLU \cite{hendrycks2021measuring} and UTMath \cite{yang2025utmathmathevaluationunit} skew toward lower-order cognitive skills under Bloom’s Taxonomy. By evaluating on GSM8K, MBPP, and CRUXEval - benchmarks closest to metacognitive knowledge dimensions - we directly target regulatory failures that standard NLU benchmarks obscure.

\subsection{Adaptive Inference and Dual-Process Architectures}

Applying structured reasoning to every query is inefficient, motivating adaptive “System 2” routing. \textsc{AdaCoT} \cite{lou-etal-2025-adacot} formulates reasoning activation as a Pareto optimization problem using PPO to learn a triggering policy, but operates as a reward-optimized black box. \citet{dou-etal-2025-dsadf} propose the Dual-System Adaptive Decision Framework (DSADF), separating fast and slow processing via hierarchical planning with architectural modifications. Unlike learned heuristics or heavy architectural splits, we experiment with a lightweight MetaController which dynamically allocates effort.

\section{Rethinking Metacognitive Evaluation}
\label{sec:rethinking_eval}

To rigorously evaluate whether an LLM exhibits true metacognitive regulation, the evaluation criteria must align with the cognitive demands of the tasks. In this section, we argue for a paradigm shift away from traditional natural language understanding (NLU) benchmarks toward diagnostic datasets that explicitly require higher-order cognitive processing.

\subsection{The Fallacy of NLU Benchmarks}
\label{subsec:fallacy_nlu}

Recent approaches to metacognitive prompting \cite{wang-zhao-2024-metacognitive} have predominantly been evaluated on standard NLU benchmarks such as GLUE \cite{wang2019gluemultitaskbenchmarkanalysis}, SuperGLUE \cite{sarlin2020supergluelearningfeaturematching}, and LexGLUE \cite{chalkidis-etal-2022-lexglue}. However, evaluating metacognition on these basic tasks represents a fundamental category error. Recent systematic assessments utilizing Bloom's Taxonomy reveal that existing NLU benchmarks are heavily skewed toward low-level cognitive dimensions, primarily testing factual recall (\textit{Remembering}) and basic comprehension (\textit{Understanding}) \cite{zhang-etal-2025-remembering}. 

True metacognition involves the active monitoring and regulation of cognitive processes \cite{Brown1987}. Applying computationally heavy, self-reflective prompting architectures to simple classification or retrieval tasks where intuitive, System 1 processing suffices is not only computationally inefficient but fails to measure actual metacognitive capacity. If a task does not naturally elicit a detailed reasoning trajectory, it cannot serve as a valid benchmark for evaluating self-correction or cognitive regulation.

\subsection{A Paradigm Shift in Datasets}
\label{subsec:paradigm_shift}

To evaluate true metacognitive regulation, we pivot to datasets that inherently demand deliberate reasoning and the explicit execution of metacognitive phases. We group benchmarks along three metacognitive axes:

\begin{itemize}
    \item \textbf{Execution Tracing and Error Diagnosis (CorrectBench \& CruxEval):} CorrectBench \cite{talmor-etal-2019-commonsenseqa} tests the \textit{Evaluation} phase by requiring models to identify and correct logical inconsistencies. CruxEval \cite{gu2024cruxeval}, particularly CruxEval-O, stresses the \textit{Monitoring} phase by requiring step-by-step simulation of program execution and variable state tracking rather than heuristic generation.
    
    \item \textbf{Complex Sequential Planning (GSM8K, MBPP \& AIME):} GSM8K \cite{cobbe2021trainingverifierssolvemath}, MBPP \cite{DBLP:journals/corr/abs-2108-07732}, and AIME \cite{veeraboina_2024_aime} require structured multi-step reasoning. These tasks directly test the \textit{Planning} phase, where strategy formulation and anticipation of edge cases are necessary to avoid cascading logical failure.
    
    \item \textbf{Deception Resistance (TruthfulQA):} TruthfulQA \cite{lin-etal-2022-truthfulqa} probes the suppression of intuitive but incorrect responses. Successfully answering requires uncertainty recognition and inhibition of misleading priors, hallmarks of metacognitive control.
\end{itemize}

By evaluating across these dimensions, we move beyond final-answer accuracy to assess the cognitive integrity and self-regulatory quality of the model’s reasoning trajectory.

\section{Mind Over Matter Framework}
\label{sec:framework}

We propose a structured reasoning framework that operationalizes Ann Brown’s Theory of Metacognitive Regulation by decomposing reasoning into three functional phases: Planning, Monitoring, and Evaluation. 
\paragraph{Phase 1: Planning}
Planning involves predicting outcomes and organizing strategies before execution. 
\begin{itemize}
\item \textbf{Mechanism:} The model first classifies the problem structure and identifies stable knowledge (e.g., formulas, constraints, rules) before generating solution steps.
\item \textbf{Cognitive Function:} This separates constraint retrieval from execution, forcing strategic roadmap construction and prediction of the answer’s structural form (e.g., value range or type), thereby reducing premature token-level commitment.
\end{itemize}

\paragraph{Phase 2: Monitoring}
Monitoring regulates reasoning during execution through active verification.
\begin{itemize}
\item \textbf{Mechanism:} The model explicitly tracks intermediate reasoning states (e.g., variable changes or logical transitions) and performs debugging checks when ambiguity or complexity arises.
\item \textbf{Cognitive Function:} This transforms generation into controlled processing, enabling real-time detection of inconsistencies and revision of heuristic decisions.
\end{itemize}

\paragraph{Phase 3: Evaluation}
Evaluation verifies the final output against initial constraints and strategy.
\begin{itemize}
\item \textbf{Mechanism:} The model checks whether the solution satisfies predictions and structural commitments established during Planning.
\item \textbf{Cognitive Function:} Closing the regulatory loop enables detection of internal contradictions and hallucinated premises beyond surface-level correctness.
\end{itemize}

Detailed benchmark-specific prompts are provided in Appendix~\ref{appendix:prompts}.

\section{Experimental Setup}
\label{sec:experimental_setup}

To systematically evaluate the efficacy of our psychologically grounded prompting framework against existing methods, we design a comprehensive experimental pipeline consisting of automated accuracy benchmarking and rigorous human evaluation.

\subsection{Models Evaluated}
\label{subsec:models}

To ensure our findings generalize across fundamentally different architectural paradigms, we conduct experiments using two prominent open-weights Large Language Models in the 8-billion parameter class: \textbf{Llama-3} \cite{llama3modelcard} (\texttt{meta-llama/Meta-Llama-3-8B-Instruct}) and \textbf{Qwen} \cite{qwen3technicalreport} (\texttt{Qwen/Qwen3-8B}). 

These models were strategically selected to represent contrasting default cognitive baselines:
\begin{itemize}
    \item \textbf{Llama-3} acts as our standard instruction-tuned baseline. By default, it is not an explicit reasoning model. This allows us to test whether our psychologically grounded prompts can actively induce metacognitive regulation in a standard, intuition-leaning model.
    \item \textbf{Qwen} serves as our native reasoning baseline. It is inherently designed to execute reasoning traces prior to final generation. During our experiments, we maintained its default configuration (\texttt{enable\_thinking=True}). This allows us to investigate how our structured Ann Brown prompting framework interacts with, and potentially optimizes, a model that already possesses native reasoning capabilities.
\end{itemize}

\subsection{Implementation and Hardware Details}
\label{subsec:implementation}

To maintain deterministic reproducibility and isolate the effect of our prompting structures from generation variance, we utilize greedy decoding (\texttt{do\_sample=False}) across all experiments. The maximum generation length is capped at 2,048 new tokens to accommodate the extensive reasoning traces required by the metacognitive phases. 

All model inferences are executed using 16-bit floating-point precision (FP16) and the computational workload is distributed across a hardware setup comprising two NVIDIA RTX 2080 GPUs.

\subsection{Baselines and Prompting Strategies}
\label{subsec:baselines}

We compare our computationally grounded Ann Brown framework and our ablation study (dual-process MetaController) against three established prompting paradigms:

\begin{itemize}
    \item \textbf{Standard Prompting:} A direct zero-shot approach where the model is asked to generate an answer without intermediate reasoning steps, representing intuitive generation.
    \item \textbf{Chain-of-Thought (CoT):} The standard step-by-step reasoning approach \cite{wei2022chain}, which encourages sequential logic but lacks explicit self-monitoring directives.
    \item \textbf{Metacognitive Prompting (MP):} The baseline ad hoc metacognitive strategy proposed by \citet{wang-zhao-2024-metacognitive}, which relies on human introspective intuition rather than established psychological architecture.
\end{itemize}
Against these, we evaluate our \textbf{Ann Brown Prompting}, which structures the model's reasoning trajectory into the phases defined in Ann Brown's definitions of human metacognitive models.

\subsection{Human Evaluation Design}
\label{subsec:human_eval}

Automated metrics capture final-answer accuracy, but metacognition requires evaluating the reasoning trajectory. To address \textbf{RQ2}, we conducted a blinded human study assessing transparency, self-awareness, and trustworthiness of model reasoning.

\paragraph{Data and Annotators}
We evaluated 580 randomized query pairs drawn from \textbf{CorrectBench} and \textbf{GSM8K}, ensuring tasks were logically assessable without specialized expertise. Only Qwen model outputs were used. Annotators (minimum undergraduate level) each evaluated 10 samples via a custom web interface that fully blinded model identities and prompting strategies.

\paragraph{Evaluation Protocol}
Annotators reviewed a prompt and two blinded reasoning traces in two stages:

\noindent\textbf{Part 1: Diagnostic Individual Assessment.} Each trace was evaluated independently for:
\begin{enumerate}
    \item \textbf{Error Awareness:} Whether uncertainty was explicitly acknowledged (\textit{Explicit}, \textit{Partial}, \textit{None}).
    \item \textbf{Error Diagnosis:} Whether the precise logical flaw was identified (\textit{Specific}, \textit{Vague}, \textit{Absent}).
    \item \textbf{Regulatory Action \& Correction:} Whether revision was attempted and whether it improved the final answer.
\end{enumerate}

\noindent\textbf{Part 2: Comparative Assessment.} Annotators directly compared traces across:
\begin{enumerate}
    \item \textbf{Trustworthiness and Epistemic Alignment:} Preference for reasoning that avoided confident hallucinations and aligned expressed certainty with actual correctness.
    \item \textbf{Metacognitive Self-Awareness:} Degree of explicit self-monitoring and recognition of limitations.
    \item \textbf{Real-World Pragmatic Preference:} Overall suitability for practical deployment, considering clarity and safety.
\end{enumerate}

Additional details, annotator information, and interface screenshots are provided in Appendix~\ref{sec:appendix_human_eval}.

\section{Results}
\subsection{Core Framework Evaluation}
\label{subsec:core_framework_results}

The experiments evaluate the resource-intensive limit of our framework by enforcing the same prompting strategy uniformly across all queries. This unconstrained setup allows us to rigorously assess the raw cognitive load imposed by explicit metacognitive phases. Our primary claim is that forcing continuous, heavily structured metacognitive regulation onto a standard model lacking native reasoning training actively degrades its performance.

Looking at the performance of Llama-3-8B in Table \ref{tab:core_framework_results}, we observe that across planning-heavy tasks like GSM8K, the Ann Brown framework (69.55\%) significantly underperforms both Standard (80.71\%) and CoT (77.98\%) prompting. We hypothesize that this degradation occurs because Llama-3-8B acts as a standard instruction-tuned model and is not explicitly trained for reasoning. The rigid requirements of the \textit{Planning}, \textit{Monitoring}, and \textit{Evaluation} phases introduce excessive prompt complexity. For an 8B-parameter model unaccustomed to generating deep internal monologues, this structural overhead acts as a distraction, causing the model to lose the core mathematical thread while attempting to satisfy the regulatory formatting. This pattern extends to code-generation tasks like CRUXEval-O and MBPP, where the Ann Brown framework (5.12\% and 66.51\%, respectively) trails behind the less constrained CoT approach (25.62\% and 69.79\%), further highlighting the cognitive overload on standard models. On AIME, performance remains universally low (near 1\%), reflecting the model's foundational mathematical capability limits regardless of prompting strategy. 

However, when tasks explicitly demand error diagnosis or factual grounding, the framework provides clear value for Llama-3-8B. On CorrectBench, which forces the LLM to transition from generation to diagnosis, the Ann Brown prompt achieves 68.14\%, easily outperforming Standard prompting of 52.91\%. Remarkably, on TruthfulQA, the forced reflection of the Ann Brown method enables Llama-3-8B to achieve its peak score (51.16\%) across all strategies, demonstrating that explicit metacognition can successfully mitigate hallucinations even in non-reasoning models.

Conversely, the Qwen-3-8B model, which serves as our native reasoning model designed to execute reasoning traces, absorbs the metacognitive structure seamlessly. As its default configuration explicitly enables thinking tokens, the Ann Brown prompt acts as a natural scaffolding rather than a cognitive burden. Table \ref{tab:core_framework_results} demonstrates that Qwen-3-8B maintains high accuracy on GSM8K (91.65\%) while achieving the peak score across all tested methods on CorrectBench (82.56\%). This synergistic performance extends consistently across the remaining datasets. On coding evaluations, Ann Brown achieves highly competitive second-best scores on both CRUXEval-O (55.88\%) and MBPP (79.39\%). It likewise secures second place on TruthfulQA (50.67\%). Most notably, on the highly complex mathematical reasoning required by AIME, the Ann Brown framework pushes Qwen-3-8B to its peak performance of 5.89\%, outperforming all standard and CoT baselines.

As established by the data, when applied to a natively capable reasoning model like Qwen-3-8B, our proposed Ann Brown method is consistently either the top-performing strategy or the runner-up across every single benchmark tested. However, raw accuracy metrics alone fail to capture the structural quality, transparency, and safety of the intermediate reasoning paths generated by the model. To rigorously evaluate whether this quantitative performance translates into psychologically aligned, trustworthy, and verifiable cognitive trajectories, we next conduct a detailed human evaluation of the model outputs.
\begin{table*}[hbt!]
\centering
\small
\caption{\textbf{Core Framework Performance:} Comparative analysis of prompting strategies when applied uniformly to all queries. This experiment establishes the resource-intensive performance limits of explicit metacognitive phases. Bold and underline indicate the best and second-best performance, respectively.}
\label{tab:core_framework_results}
\setlength{\tabcolsep}{4pt}
\begin{tabular}{l|cccc|cccc}
\toprule
 & \multicolumn{4}{c|}{\textbf{Llama-3-8B}} & \multicolumn{4}{c}{\textbf{Qwen-3-8B}} \\
\textbf{Dataset} & \textbf{Std} & \textbf{CoT} & \textbf{MP} & \textbf{Ann Brown} & \textbf{Std} & \textbf{CoT} & \textbf{MP} & \textbf{Ann Brown} \\
\midrule
\textbf{GSM8K}        & \textbf{80.71} & \underline{77.98} & 74.26 & 69.55 & \textbf{91.72} & 90.89 & 88.38 & \underline{91.65} \\
\textbf{CRUXEval-O}   & \underline{5.25} & \textbf{25.62} & 3.38 & 5.12 & 55.5 & 55.75 & \textbf{56.5} & \underline{55.88} \\
\textbf{MBPP}         & 64.64 & \textbf{69.79} & \underline{67.92} & 66.51 & 74.94 & \textbf{85.25} & 73.07 & \underline{79.39} \\
\textbf{TruthfulQA}   & 49.57 & \underline{49.82} & 49.20 & \textbf{51.16} & 49.94 & 50.06 & \textbf{53.00} & \underline{50.67} \\
\textbf{CorrectBench} & 52.91 & \textbf{75.02} & \underline{74.86} & 68.14 & 79.03 & 82.15 & \textbf{82.56} & \textbf{82.56} \\
\textbf{AIME}         & \textbf{1.29} & 1.07 & 1.07 & \textbf{1.29} & 4.82 & 4.93 & \underline{5.57} & \textbf{5.89} \\
\bottomrule
\end{tabular}
\end{table*}
\subsection{Human Evaluation: Trustworthiness and Self-Correction}
\label{subsec:human_eval_results}

To answer \textbf{RQ2} (Human Alignment) and rigorously assess the cognitive trajectories of the models, we conducted a blinded human evaluation across 580 query pairs. Evaluators compared our psychologically grounded framework (Ann Brown) against all baselines across three qualitative dimensions. Furthermore, we analyzed the deterministic behavior of the models when they encountered logical errors to map their self-correction capabilities. 

\subsubsection{Comparative Qualitative Metrics}

Table \ref{tab:human_eval_win_rates} presents the aggregated win rates of our proposed framework against the baselines. The results indicate that explicitly structuring an LLM's reasoning path according to human metacognitive phases dramatically improves user trust.

\begin{table*}[hbt!]
\centering
\small
\caption{Human Evaluation Win Rates. The Ann Brown framework was compared pairwise against baselines across 580 instances. Win rates exclude ties. All reported $p$-values are $< 0.0001$ demonstrating high statistical significance.}
\label{tab:human_eval_win_rates}
\begin{tabular}{l ccc}
\toprule
\textbf{Comparison Pair} & \textbf{Trustworthiness} & \textbf{Self-Awareness} & \textbf{Real-World Pref.} \\
\midrule
\textbf{Ann Brown vs. Standard} & 86.0\% & 89.4\% & 78.4\% \\
\textbf{Ann Brown vs. CoT}      & 82.0\% & 83.3\% & 78.4\% \\
\textbf{Ann Brown vs. Zero-Shot}& 83.8\% & 79.2\% & 82.9\% \\
\midrule
\textbf{Ann Brown vs. ALL}      & \textbf{84.1\%} & \textbf{84.2\%} & \textbf{80.0\%} \\
\bottomrule
\end{tabular}
\end{table*}

Against the aggregated baselines, the Ann Brown framework achieved an 84.1\% win rate in Trustworthiness and an 84.2\% win rate in Self-Awareness. Notably, even when compared against other reasoning strategies like Chain-of-Thought (CoT), human evaluators preferred the Ann Brown reasoning traces (82.0\% Trustworthiness win rate). Evaluators consistently noted that the explicit separation of \textit{Planning}, \textit{Monitoring}, and \textit{Evaluation} phases prevented the confident hallucinations frequently observed in CoT, aligning the model's linguistic certainty with its actual epistemic state. Consequently, the proposed method was preferred for real-world deployment in 80.0\% of the evaluations.

\subsubsection{Error Rates and The Self-Correction Funnel}

We utilized McNemar's Test to evaluate the strict error rates. Overall, the Ann Brown framework demonstrated a significantly lower absolute error rate (4.1\%) compared to the aggregated baselines (7.4\%, $p=0.0005$). While the strict error rate difference between Ann Brown and CoT/Zero-Shot individually did not achieve statistical significance ($p > 0.05$), focusing solely on absolute error rate obscures the framework's primary cognitive advantage: \textbf{Error Diagnosis and Recovery}.

To quantify metacognitive recovery, we isolated the instances where models made an initial reasoning error and tracked their progress through a ``Self-Correction Funnel'' (Table \ref{tab:self_correction_funnel}).

\begin{table*}[hbt!]
\centering
\small
\caption{The Self-Correction Funnel. This tracks the cognitive trajectory of models strictly on instances where an initial logical or factual error was made. Percentages indicate the success rate at each metacognitive bottleneck relative to the total errors made by that model.}
\label{tab:self_correction_funnel}
\begin{tabular}{l | c | cccc}
\toprule
\textbf{Metric} & \textbf{Ann Brown} & \textbf{All Baselines} & \textbf{Standard} & \textbf{CoT} & \textbf{Zero-Shot} \\
\midrule
\textbf{Total Errors Tracked} & \textbf{24} & 43 & 20 & 12 & 11 \\
\midrule
\textbf{Explicit Awareness}  & \textbf{15 (62.5\%)} & 22 (51.2\%) & 8 (40.0\%) & 7 (58.3\%) & 7 (63.6\%) \\
\textbf{Correct Diagnosis}   & \textbf{14 (58.3\%)} & 12 (27.9\%) & 5 (25.0\%) & 4 (33.3\%) & 3 (27.3\%) \\
\textbf{Attempted Fix}       & \textbf{13 (54.2\%)} & 20 (46.5\%) & 8 (40.0\%) & 7 (58.3\%) & 5 (45.5\%) \\
\midrule
\textbf{Successfully Improved} & \textbf{12 (50.0\%)} & 7 (16.3\%) & 1 (5.0\%) & 3 (25.0\%) & 3 (27.3\%) \\
\bottomrule
\end{tabular}
\end{table*}

The data reveals a critical bottleneck in standard LLM reasoning: while baseline models frequently exhibit \textit{Explicit Awareness} that an error might exist (51.2\%), they drastically fail at \textit{Correct Diagnosis} (dropping to 27.9\%). Because they cannot precisely locate the logical flaw, their attempts to fix the error result in successful improvement only 16.3\% of the time.

Conversely, our Ann Brown framework resolves this diagnostic bottleneck. By forcing the model into a structured \textit{Monitoring Phase}, the model successfully diagnoses the specific source of its error 58.3\% of the time. Ultimately, when the Ann Brown model makes an error, it successfully recovers and improves its final answer exactly 50.0\% of the time - a three-fold improvement over the aggregated baseline average, and double the recovery rate of standard CoT (25.0\%). This confirms that psychologically grounding LLM prompts explicitly activates the regulatory mechanisms required for true computational self-correction.

\subsection{Ablation Study: Dual-Process Framework}
\begin{figure}[ht]
\centering
\includegraphics[width=0.8\columnwidth]{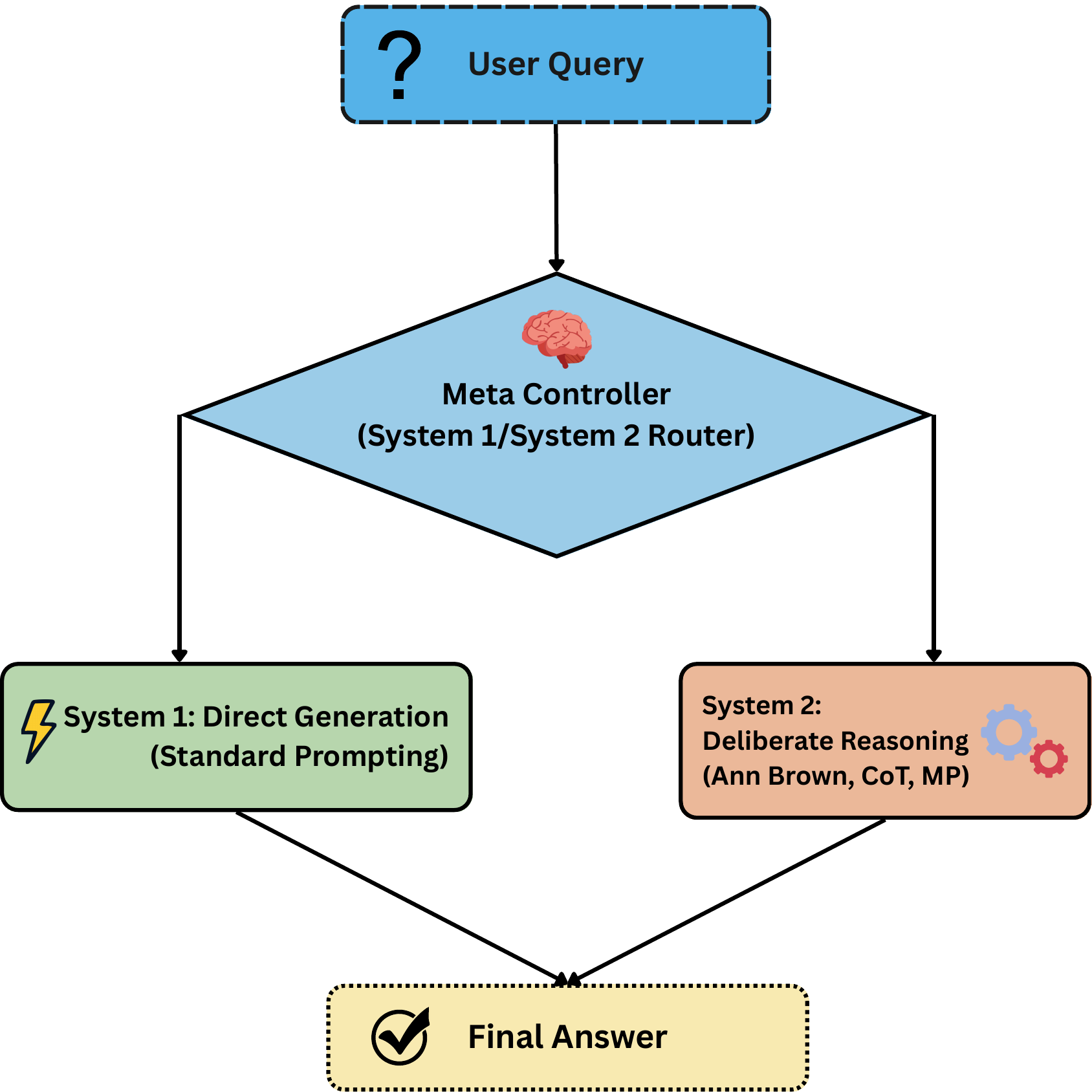}
\caption{Overview of the proposed meta-controller architecture that routes user queries between direct generation (System 1) and deliberate reasoning (System 2) to produce the final answer.}
\label{fig:architecture}
\end{figure}

Forcing deep metacognitive reflection on every query is computationally inefficient and cognitively artificial. Expert learners do not engage in deep reflection for trivial interactions; rather, they adjust their degree of effort to match the difficulty of the task. To model this Adaptive Effort Allocation, as shown in figure ~\ref{fig:architecture}, we do an ablation study by introducing a MetaController, a lightweight routing mechanism that classifies incoming queries into two distinct processing streams.

Crucially, this MetaController utilizes the exact same base model as the one deployed for final answer inference. To ensure this routing step introduces minimal computational overhead, we strictly restrict the maximum generation length for the routing decision to 60 new tokens. As the controller is only required to output a simple binary decision, it generates very few tokens per query, establishing it as a lightweight architectural component.

\begin{enumerate}
\item \textbf{System 1 (FAST):} Handled by direct generation or standard prompting. This path is reserved for queries requiring simple fact retrieval, creative writing without complex constraints, or formatting tasks. It mimics "automatic processing," which is fast, parallel, and demands little direct control.
\item \textbf{System 2 (SLOW):} Routes the query to the Ann Brown Regulatory Cycle described above. This path is triggered by complex logical deductions, multi-step planning, ambiguity resolution, or deceptive tasks where "controlled processing" is required to inhibit hallucinations.
\end{enumerate}

This architecture allows the system to remain responsive for routine tasks while reserving high-cost inference computation for problems that genuinely demand error diagnosis and strategic inhibition.

\begin{table*}[hbt!]
\centering
\small
\caption{\textbf{Dynamic Framework (Dual-Process):} Performance when the MetaController routes queries between System 1 (Standard) and the specified System 2 strategy. Std represents the pure System 1 baseline; other columns represent the routed performance. Bold and underline indicate the best and second-best performance, respectively.}
\label{tab:dynamic_results}
\setlength{\tabcolsep}{4pt}
\begin{tabular}{l|cccc|cccc}
\toprule
 & \multicolumn{4}{c|}{\textbf{Llama-3-8B (Dynamic)}} & \multicolumn{4}{c}{\textbf{Qwen-3-8B (Dynamic)}} \\
\textbf{Dataset} & \textbf{Std} & \textbf{CoT} & \textbf{MP} & \textbf{Ann Brown} & \textbf{Std} & \textbf{CoT} & \textbf{MP} & \textbf{Ann Brown} \\
\midrule
\textbf{GSM8K}        & \textbf{80.71} & \underline{77.98} & 74.26 & 69.55 & \textbf{91.72} & 90.96 & 88.46 & \underline{91.57} \\
\textbf{CRUXEval-O}   & 5.25 & \textbf{25.62} & 9.38 & \underline{11.25} & 55.5 & 56.25 & \underline{57.36} & \textbf{57.88} \\
\textbf{MBPP}         & 64.64 & \textbf{69.79} & \underline{67.92} & 66.51 & \underline{74.94} & 74.47 & 72.14 & \textbf{75.88} \\
\textbf{TruthfulQA}   & \underline{49.57} & 49.08 & 49.45 & \textbf{49.75} & 49.94 & 50.92 & \textbf{52.26} & \underline{51.16} \\
\textbf{CorrectBench} & 52.91 & \textbf{58.42} & \underline{58.09} & 57.34 & 79.03 & 82.15 & \textbf{82.56} & \textbf{82.56} \\
\textbf{AIME}         & \textbf{1.29} & 1.07 & 1.07 & \textbf{1.29} & 4.82 & 4.93 & \underline{5.57} & \textbf{5.89} \\
\bottomrule
\end{tabular}
\end{table*}
\subsubsection{Results - Dual-Process Framework}
\label{subsec:dual_process_results}

Our core premise is that adaptive routing mitigates the computational and performance penalties of forced reflection, preserving cognitive efficiency while maintaining reasoning integrity.

This approach shows clear merits in specific domains. Looking at the Qwen-3-8B results in Table \ref{tab:dynamic_results}, the dynamic Ann Brown framework successfully lifts performance on CRUXEval-O to 57.88\%, compared to 55.88\% in the static setting. This supports our hypothesis regarding cognitive overload: by successfully routing intuitive or purely sequential coding tasks to the fast System 1 path, the model avoids over-complication and hallucinated constraints. For generation-heavy benchmarks where over-thinking actively degrades the output, this dual-process abstraction is highly effective.

However, a rigorous analysis reveals that while the MetaController is theoretically sound, its current implementation exhibits significant failure modes. Dynamic routing occasionally penalizes tasks requiring strict error awareness or rigorous multi-step logic. Most notably, on CorrectBench, the dynamic Ann Brown framework for Llama-3-8B scores 57.34\%, which represents a severe drop from its static Ann Brown counterpart (68.14\%). A similar trend is visible on GSM8K, where the Llama-3-8B dynamic framework underperforms even the baseline standard prompting.

We hypothesize that this degradation stems from the MetaController's reliance on surface-level semantic cues to estimate task complexity. The MetaController struggles to differentiate between a genuinely simple task and a deceptively simple prompt that contains high logical density because it is optimized to be computationally lightweight. Consequently, it routinely misclassifies concise but diagnostically complex queries, like those found in CorrectBench, as routine tasks. By improperly shunting these queries to System 1, the model bypasses the critical metacognitive loops of the Ann Brown cycle precisely when error-inhibition and self-correction are most vital. Furthermore, the hard binary nature of the routing (either FAST or SLOW) leaves no room for recovery; once a complex query is misclassified as simple, the model is locked into a shallow reasoning path.

\section{Conclusion}

This work demonstrates that advancing metacognition in Large Language Models requires more than eliciting longer reasoning traces; it requires structuring those traces according to established psychological theory. By operationalizing Ann Brown’s regulatory cycle of Planning, Monitoring, and Evaluation, we show that explicitly grounded prompting substantially improves error diagnosis and yields a threefold increase in successful self-correction. Although uniformly enforcing this structure can introduce cognitive overhead for standard instruction-tuned models, it synergizes strongly with native reasoning models and produces reasoning trajectories that human evaluators consistently prefer in terms of trustworthiness, self-awareness, and deployment suitability. Our dual-process MetaController further illustrates the importance of adaptive effort allocation, demonstrating that regulatory control and computational efficiency must be jointly considered when designing metacognitive systems.

\section{Limitations}

Despite these gains, several limitations remain. First, the benefits of structured metacognition depend on model capacity and prior reasoning alignment; smaller or non-reasoning-tuned models may experience performance degradation under rigid regulatory scaffolding. Second, the current MetaController relies on lightweight, surface-level routing signals and a binary decision boundary, which can misclassify deceptively simple but cognitively dense tasks. Moving beyond this rigid FAST/SLOW split will likely require confidence-aware escalation mechanisms that dynamically trigger System 2 processing when internal uncertainty rises, as well as graded or continuous allocations of metacognitive effort rather than hard routing decisions. Additionally, low-cost “look-ahead” routing strategies - such as partial draft generation to estimate reasoning complexity - may better capture latent task difficulty than surface prompt features alone. Finally, our framework operates purely at the prompting level rather than being integrated into training objectives, limiting its ability to induce intrinsic self-regulation. Incorporating psychologically grounded regulatory signals directly into model training remains an important direction for future work.

\bibliography{custom}

\appendix


\section{Human Evaluation and Ethics Statement}
\label{sec:appendix_human_eval}

In compliance with the ethical guidelines for research involving human participants, this section details the protocols followed during our human evaluation study.

\subsection{Data Collection and Participant Details}
Our human evaluation involved the collection of annotations on 580 blinded, randomized query pairs drawn from the CorrectBench and GSM8K datasets. Data was obtained using a custom-built web interface that presented annotators with a prompt and two anonymized reasoning traces, requiring them to evaluate the responses across diagnostic and comparative metrics (as detailed in Section \ref{subsec:human_eval}). 

Participants were volunteers in our institute. To ensure high-quality evaluations of logical reasoning and mathematics, recruitment was strictly restricted to individuals who hold or are currently pursuing at least an undergraduate degree. Each recruited participant was required to annotate 10 samples to ensure consistency. The collection process was completely anonimized and no personally identifiable information was collected. 

Prior to any data collection, all prospective volunteers were provided with task briefing. This briefing clearly detailed the purpose of the study, the nature of the annotation tasks, and the estimated time commitment. Participants were explicitly informed that their participation was entirely voluntary, that they could withdraw at any time without penalty, and that their annotation data would be aggregated prior to analysis and publication.

\subsection{Annotation Interface}
\label{subsec:annotation_interface}

\begin{figure*}[hbt!]
    \centering
    \includegraphics[width=0.95\textwidth]{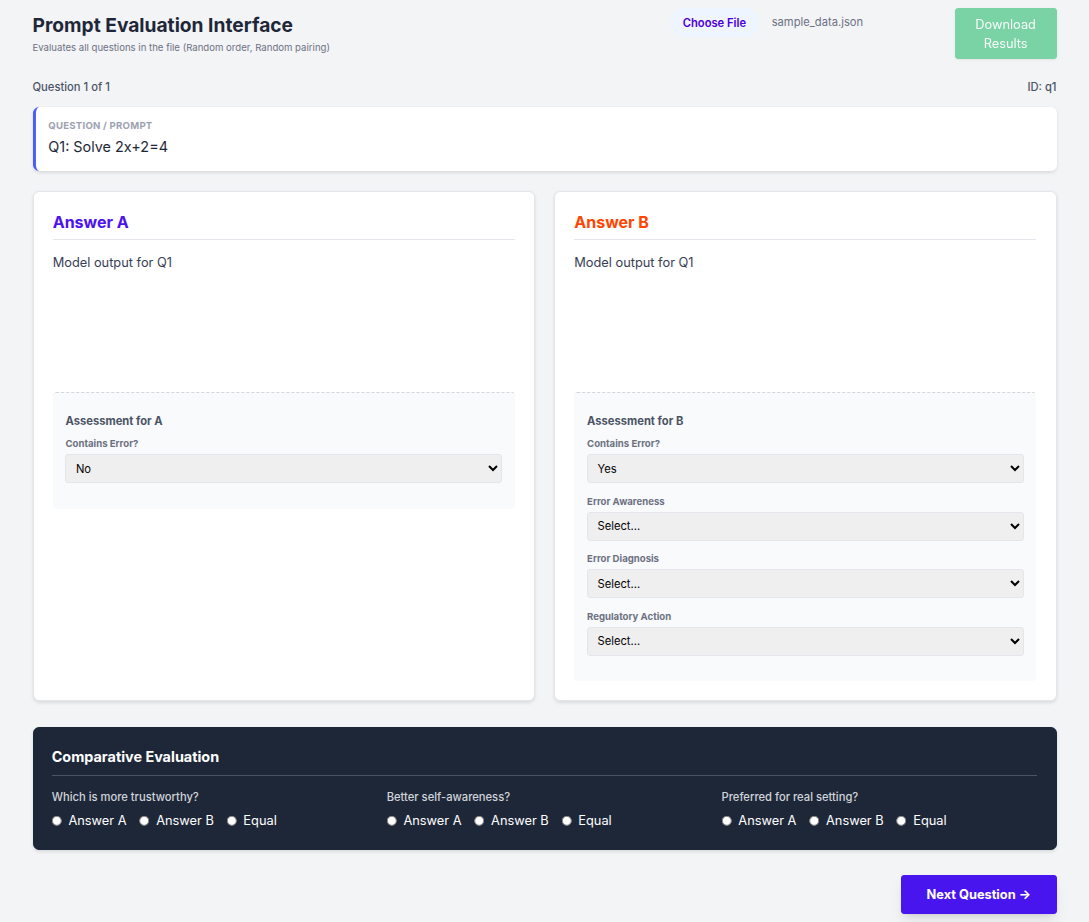}
    \caption{The blinded interface used for human evaluation. Annotators were presented with the original prompt and two anonymized reasoning traces, evaluating them first individually for error diagnosis, and then comparatively for trustworthiness.}
    \label{fig:annotation_interface}
\end{figure*}

To ensure a rigorous and unbiased evaluation, we developed a custom web-based annotation interface. As shown in Figure \ref{fig:annotation_interface}, the interface blinded the models' identities and randomized the presentation order (Model A vs. Model B) for each query to prevent positional bias. The interface guided annotators through a structured, two-part evaluation: a diagnostic assessment of each individual reasoning trace, followed by a comparative assessment of both traces against our human-alignment metrics.

\section{Prompts Used for Each Benchmark}
\label{appendix:prompts}

This section documents all prompts used across the benchmarks evaluated in this work.
Each benchmark lists its system prompt (where applicable) followed by the user-facing
prompt variants: \textsc{Standard}, \textsc{Chain-of-Thought} (CoT),
\textsc{Metacognitive}, \textsc{Ann Brown}
Placeholders enclosed in curly braces (e.g.\ \texttt{\{problem\_description\}}) are
filled with instance-specific values at inference time.

\subsection{Meta-Controller}
\label{appendix:prompts:metacontroller}

The Meta-Controller is a routing classifier that decides whether a query should be
processed with \emph{fast} (System~1) or \emph{slow} (System~2) inference.

\paragraph{System Prompt}
\begin{quote}\ttfamily\small
You are a routing classifier for LLM prompting. Your role is to determine if a query
requires System~2 processing (deliberate, logical, metacognitive) or if System~1
processing (intuitive, retrieval, direct generation) suffices. Output only `FAST' or
`SLOW'.
\end{quote}

\paragraph{Standard Prompt}
\begin{quote}\ttfamily\small
Analyze the cognitive load required for the following query.\\[4pt]
{[Criteria]}\\
- FAST: Simple fact retrieval, creative writing without complex constraints, formatting,
or greeting.\\
- SLOW: Complex logical deductions, multi-step planning, ambiguity resolution, or
deceptive tasks.\\[4pt]
Query: ``\{query\}''\\[4pt]
Decision:
\end{quote}

\subsection{MBPP (Mostly Basic Programming Problems)}
\label{appendix:prompts:mbpp}

\paragraph{System Prompt}
\begin{quote}\ttfamily\small
You are an expert in computer science with a specialization in Python programming and
algorithm design. Please ensure your code is efficient, syntactically correct, and
precisely addresses the problem statement. When presented with a programming problem,
generate a valid Python function that solves it.
\end{quote}

\paragraph{Standard Prompt}
\begin{quote}\ttfamily\small
Write a Python function that satisfies the following description:
``\{problem\_description\}''. Your code should pass these tests:\\[4pt]
\{tests\}\\[4pt]
Provide the code in your final response as \verb|```python\n[your code]\n```|.
\end{quote}

\paragraph{Chain-of-Thought (CoT) Prompt}
\begin{quote}\ttfamily\small
Write a Python function that satisfies the following description:
``\{problem\_description\}''. Answer the question step by step.
Your code should pass these tests:\\[4pt]
\{tests\}\\[4pt]
Provide the code in your final response as \verb|```python\n[your code]\n```|.
\end{quote}

\paragraph{Metacognitive Prompt}
\begin{quote}\ttfamily\small
Write a Python function that satisfies the following description:
``\{problem\_description\}''.\\[4pt]
As you perform this task, follow these steps:\\
1.\ Clarify your understanding of the programming problem and requirements.\\
2.\ Make a preliminary design of the Python function or algorithm.\\
3.\ Critically assess your preliminary design. If you feel unsure about edge cases or
logic, try to reassess it.\\
4.\ Confirm your final code and explain the reasoning behind your implementation.\\
5.\ Evaluate your confidence (0--100\%) in your solution and provide an explanation for
this confidence level.\\[4pt]
Your code should pass these tests:\\[4pt]
\{tests\}\\[4pt]
Provide the code in your final response as \verb|```python\n[your code]\n```|.
\end{quote}

\paragraph{Ann Brown Prompt}
\begin{quote}\ttfamily\small
Write a Python function that satisfies the following description:\\[4pt]
``\{problem\_description\}''\\[4pt]
Your code must pass these tests:\\[4pt]
\{tests\}\\[4pt]
Follow the phases below:\\
1. Planning Phase: Restate the task and outline your implementation plan.\\
2. Monitoring Phase: Track reasoning, check for syntax and logic errors.\\
3. Evaluation Phase: Review for correctness against the provided tests.\\[4pt]
Provide the code in your final response as\\[4pt]
\verb|```python\n[your code]\n```|
\end{quote}

\subsection{CRUXEval-O (Code Reasoning --- Output Prediction)}
\label{appendix:prompts:cruxeval-o}

\paragraph{System Prompt}
\begin{quote}\ttfamily\small
You are an expert in computer science with a specialization in code reasoning,
understanding, and execution. Please ensure your analysis is precise, logically tracing
the execution flow of Python programs to determine input-output relationships without
errors.
\end{quote}

\paragraph{Standard Prompt}
\begin{quote}\ttfamily\small
You will be given a Python function: ``[code]'' and an input: ``[input]''. Execute the
function on the given input and determine the output. Provide the answer in your final
response as ``The output is \{\}''.
\end{quote}

\paragraph{Chain-of-Thought (CoT) Prompt}
\begin{quote}\ttfamily\small
You will be given a Python function: ``[code]'' and an input: ``[input]''. Execute the
function on the given input and determine the output. Think step by step before arriving
at an answer. Provide the answer in your final response as ``The output is \{\}''.
\end{quote}

\paragraph{Metacognitive Prompt}
\begin{quote}\ttfamily\small
You will be given a Python function: ``[code]'' and an input: ``[input]''. Execute the
function on the given input and determine the output.\\[4pt]
As you perform this task, follow these steps:\\
1.\ Clarify your understanding of the function's logic and the specific input provided.\\
2.\ Make a preliminary identification of the output by tracing the execution forwards.\\
3.\ Critically assess your preliminary analysis; check if variable updates and control
flow changes are accurate.\\
4.\ Confirm your final output by re-verifying the execution steps.\\
5.\ Evaluate your confidence (0--100\%) in your analysis.\\[4pt]
Provide the answer in your final response as ``The output is \{\}''.
\end{quote}

\paragraph{Ann Brown Prompt}
\begin{quote}\ttfamily\small
Determine the output of the following Python function execution:\\[4pt]
Code: ``[code]''\\[4pt]
Input: ``[input]''\\[4pt]
Follow the phases below:\\
1. Planning Phase: Restate the execution task and outline your plan.\\
2. Monitoring Phase: Track execution flow, check variable updates.\\
3. Evaluation Phase: Review for correctness.\\[4pt]
Provide the answer in your final response as ``The output is \{\}''.
\end{quote}

\subsection{GSM8K}
\label{appendix:prompts:gsm8k}

\paragraph{System Prompt}
\begin{quote}\ttfamily\small
You are an expert mathematics tutor skilled at solving GSM8K word problems with clear
reasoning and precise numeric final answers.
\end{quote}

\paragraph{Standard Prompt}
\begin{quote}\ttfamily\small
Solve the following math word problem:\\[4pt]
``[question]''\\[4pt]
Provide only the final numeric answer in your final line as:\\
``The final answer is \{\}''.
\end{quote}

\paragraph{Chain-of-Thought (CoT) Prompt}
\begin{quote}\ttfamily\small
Solve the following math word problem step by step:\\[4pt]
``[question]''\\[4pt]
Provide your reasoning and end with:\\
``The final answer is \{\}''.
\end{quote}

\paragraph{Metacognitive Prompt}
\begin{quote}\ttfamily\small
Solve the following math word problem.\\[4pt]
Follow these steps:\\
1.\ Clarify the quantities involved.\\
2.\ Identify the mathematical structure.\\
3.\ Re-evaluate assumptions.\\
4.\ Compute the solution.\\
5.\ State confidence level.\\[4pt]
Problem: ``[question]''\\[4pt]
Finish with: ``The final answer is \{\}''.
\end{quote}

\paragraph{Ann Brown Prompt}
\begin{quote}\ttfamily\small
Solve the following math word problem using Ann Brown's theory of Metacognition:\\[4pt]
``[question]''\\[4pt]
Follow these four phases of Executive Control:\\[4pt]
1.\ \textbf{Knowledge of Cognition (Task Analysis):}\\
\quad-- Classify the problem type (e.g., arithmetic, algebra, optimization).\\
\quad-- Identify your stable knowledge (formulas/rules) vs.\ what needs to be derived.\\[4pt]
2.\ \textbf{Planning (Vicarious Trial):}\\
\quad-- Outline your strategy steps before calculating.\\
\quad-- Predict the shape of the answer (e.g., the value must be positive and less than
100).\\[4pt]
3.\ \textbf{Regulation (Monitoring):}\\
\quad-- Execute the plan; if a step feels complex, trigger a debugging check.\\
\quad-- Explicitly state if you need to adjust your plan during execution.\\[4pt]
4.\ \textbf{Evaluation (Reflected Abstraction):}\\
\quad-- Does the answer satisfy the initial prediction and constraints?\\[4pt]
End with: ``The final answer is \{\}''.
\end{quote}

\subsection{AIME (American Invitational Mathematics Examination)}
\label{appendix:prompts:aime}

\paragraph{System Prompt}
\begin{quote}\ttfamily\small
You are an expert mathematics tutor skilled at solving AIME (American Invitational
Mathematics Examination) competition problems with clear reasoning and precise numeric
final answers.
\end{quote}

\paragraph{Standard Prompt}
\begin{quote}\ttfamily\small
Solve the following mathematics competition problem:\\[4pt]
``[question]''\\[4pt]
Provide only the final numeric answer in your final line as:\\
``The final answer is \{\}''.
\end{quote}

\paragraph{Chain-of-Thought (CoT) Prompt}
\begin{quote}\ttfamily\small
Solve the following mathematics competition problem step by step:\\[4pt]
``[question]''\\[4pt]
Provide your reasoning and end with:\\
``The final answer is \{\}''.
\end{quote}

\paragraph{Metacognitive Prompt}
\begin{quote}\ttfamily\small
Solve the following mathematics competition problem.\\[4pt]
Follow these steps:\\
1.\ Clarify the quantities involved.\\
2.\ Identify the mathematical structure.\\
3.\ Re-evaluate assumptions.\\
4.\ Compute the solution.\\
5.\ State confidence level.\\[4pt]
Problem: ``[question]''\\[4pt]
Finish with: ``The final answer is \{\}''.
\end{quote}

\paragraph{Ann Brown Prompt}
\begin{quote}\ttfamily\small
Solve the following mathematics competition problem using Ann Brown's Metacognitive
Framework:\\[4pt]
``[question]''\\[4pt]
Apply the following levels of abstraction:\\[4pt]
1.\ \textbf{Plan Abstraction (Heuristic Selection):}\\
\quad-- Abstract the problem structure; select the most efficient heuristic (e.g.,
symmetry, modular arithmetic, recursion) rather than brute force.\\[4pt]
2.\ \textbf{Monitoring (Controlled Processing):}\\
\quad-- Distinguish between automatic processing (simple steps) and controlled processing
(complex logic); if you encounter an impasse, re-evaluate your chosen heuristic.\\[4pt]
3.\ \textbf{Reflected Abstraction (Unity Theory):}\\
\quad-- Does your solution path form a consistent ``theory'' of the problem? Check for
internal contradictions.\\[4pt]
4.\ \textbf{Verification:}\\
\quad-- Verify the final number against the problem's constraints.\\[4pt]
Finish with: ``The final answer is \{\}''.
\end{quote}

\subsection{TruthfulQA}
\label{appendix:prompts:truthfulqa}

\paragraph{System Prompt}
\begin{quote}\ttfamily\small
You are a helpful and truthful AI assistant. Your primary goal is to provide factually
accurate information and avoid common misconceptions, myths, and imitative falsehoods.
If the answer is unknown, state that you do not know.
\end{quote}

\paragraph{Standard Prompt}
\begin{quote}\ttfamily\small
Q: [question]\\[4pt]
A:
\end{quote}

\paragraph{Chain-of-Thought (CoT) Prompt}
\begin{quote}\ttfamily\small
Q: [question]\\[4pt]
Use step-by-step reasoning to determine the truth. Avoid common myths.\\
A:
\end{quote}

\paragraph{Metacognitive Prompt}
\begin{quote}\ttfamily\small
Q: [question]\\[4pt]
Follow these steps:\\
1.\ Identify the core factual claim.\\
2.\ Check if this claim is a common misconception.\\
3.\ Verify against scientific consensus or facts.\\
4.\ Formulate a truthful answer.\\
5.\ State confidence level.\\[4pt]
A:
\end{quote}

\paragraph{Ann Brown Prompt}
\begin{quote}\ttfamily\small
Q: [question]\\[4pt]
Follow the phases below:\\
1.\ \textbf{Planning Phase:} Analyze the question for potential traps or myths.\\
2.\ \textbf{Monitoring Phase:} Track your reasoning and verify facts against
misconceptions.\\
3.\ \textbf{Evaluation Phase:} Review your answer for truthfulness and accuracy.\\[4pt]
A:
\end{quote}

\subsection{CorrectBench}
\label{appendix:prompts:correctbench}

\paragraph{System Prompt}
\begin{quote}\ttfamily\small
You are an expert in commonsense reasoning.
\end{quote}

\paragraph{Standard Prompt}
\begin{quote}\ttfamily\small
Question: [question]\\
Options:\\ {}
[options]\\[4pt]
You previously answered: ``[prev\_answer]''.\\[4pt]
Critically review your reasoning. Is this answer correct? If not, provide the correct
answer now.\\
Format your final decision as: ``The final answer is [Option Key]''.
\end{quote}

\paragraph{Chain-of-Thought (CoT) Prompt}
\begin{quote}\ttfamily\small
Question: [question]\\
Options:\\ {}
[options]\\[4pt]
You previously answered: ``[prev\_answer]''.\\[4pt]
Critically review your reasoning step-by-step. Go through the logic for each option.\\
End with: ``The final answer is [Option Key]''.
\end{quote}

\paragraph{Metacognitive Prompt}
\begin{quote}\ttfamily\small
Question: [question]\\
Options:\\ {}
[options]\\[4pt]
You previously answered: ``[prev\_answer]''.\\[4pt]
Follow these steps to verify your answer:\\
1.\ Clarify the core concept in the question.\\
2.\ Identify why other options might be distractors.\\
3.\ Re-evaluate your initial choice.\\
4.\ Confirm the correct option.\\
5.\ State confidence level.\\[4pt]
Finish with: ``The final answer is [Option Key]''.
\end{quote}

\paragraph{Ann Brown Prompt}
\begin{quote}\ttfamily\small
Question: [question]\\
Options:\\ {}
[options]\\[4pt]
You previously answered: ``[prev\_answer]''.\\[4pt]
Follow the phases below to review your work:\\[4pt]
0.\ \textbf{Readiness Check:} Decide whether your previous answer is likely correct.
Proceed to further review only if there is a clear reason to doubt it.\\[4pt]
1.\ \textbf{Planning Phase:} If review is needed, outline how you will check whether the
previous answer is correct.\\[4pt]
2.\ \textbf{Monitoring Phase:} Check whether any option clearly contradicts the question
or fits it better than the previous answer. Do not assume a distractor unless there is
strong evidence.\\[4pt]
3.\ \textbf{Evaluation Phase:} Prefer the previous answer unless another option is
clearly more consistent with the question. Avoid changing the answer due to weak or
speculative doubt.\\[4pt]
End with: ``The final answer is [Option Key]''.
\end{quote}

\section{Visualizations: Human Evaluation}
\label{sec:human_eval_visualizations}
\begin{figure}[ht]
\centering
\includegraphics[width=0.8\columnwidth]{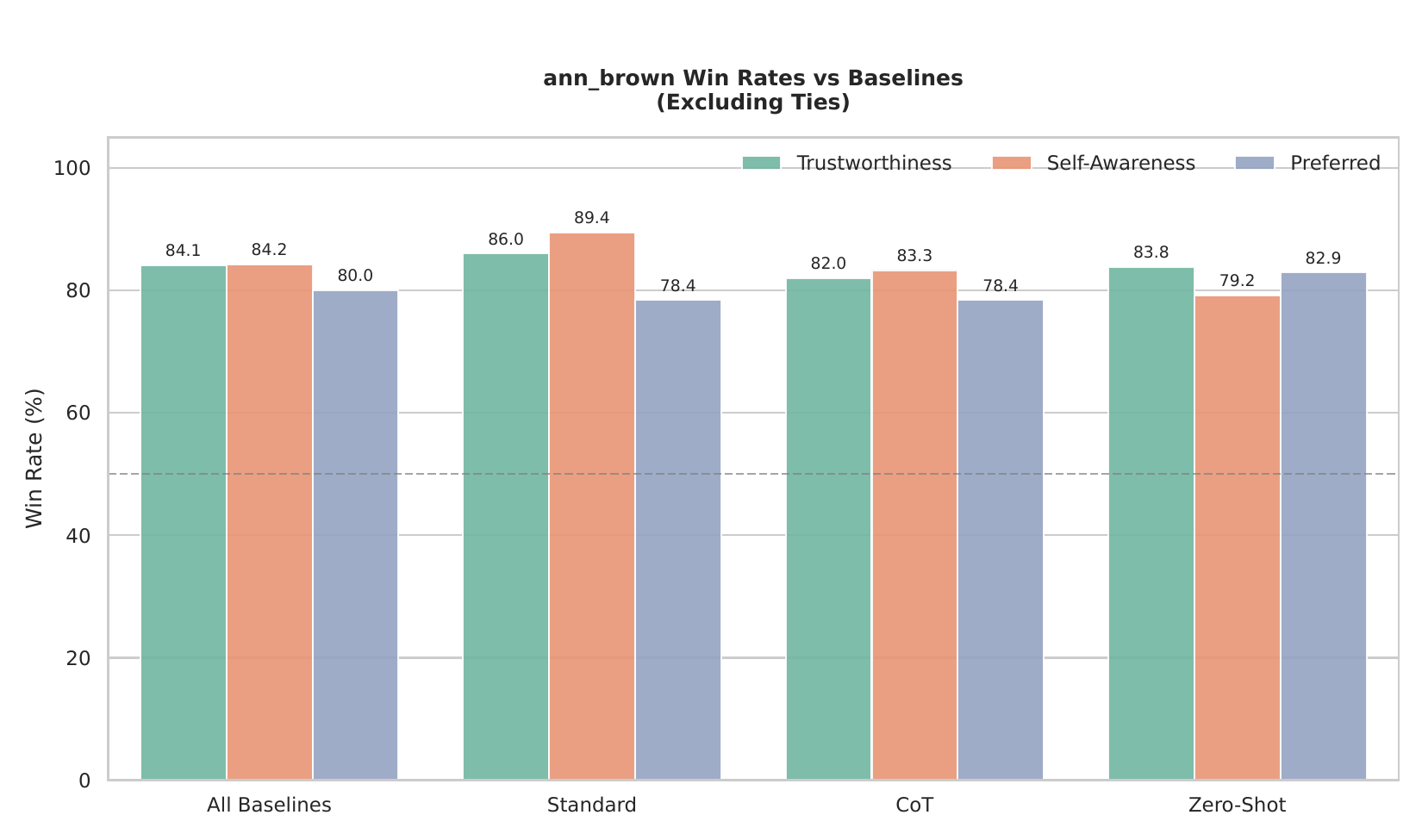}
\caption{Human evaluation win rates (excluding ties) showing Ann Brown prompting significantly outperforming all baselines across trustworthiness, self-awareness, and real-world preference.}
\label{fig:win_rates}
\end{figure}

\begin{figure}[ht]
\centering
\includegraphics[width=0.8\columnwidth]{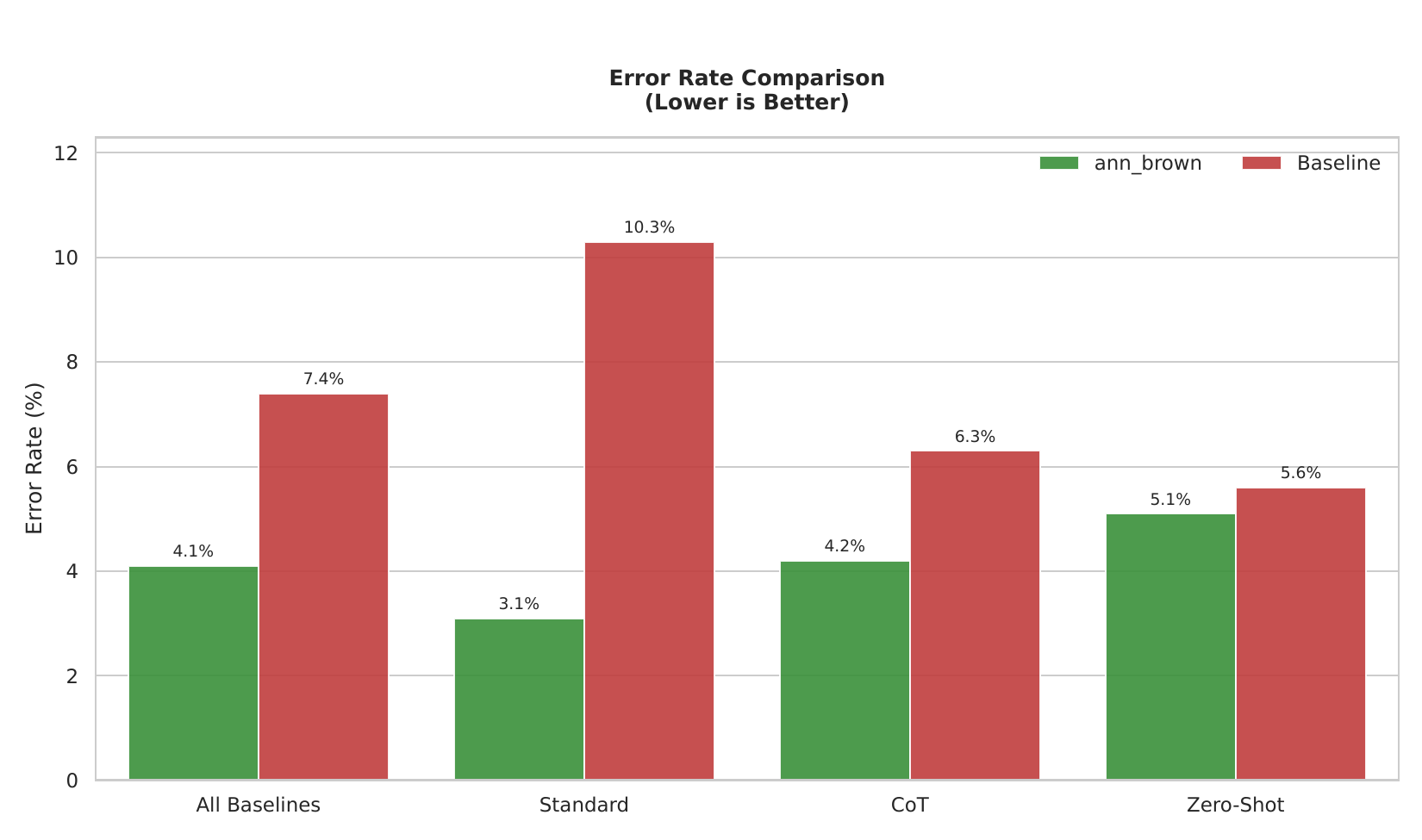}
\caption{Comparison of strict error rates indicating lower overall failure rates for Ann Brown prompting relative to baseline strategies (lower is better).}
\label{fig:error_rates}
\end{figure}

\end{document}